# The unreasonable effectiveness of AI CADe polyp detectors to generalize to new countries

*Joel Shor[1], *Hiro-o Yamano[4], *Daisuke Tsurumaru[3], *Yotami Intrator[1], Hiroki Kayama[2], Joe Ledsam[2], Atsushi Hamabe[5], Koji Ando[3], Mitsuhiko Ota[3], Haruei Ogino[5], Hiroshi Nakase[4], Kaho Kobayashi[2], **Eiji Oki[3], **Roman Goldenberg[1], **Ehud Rivlin[1], **Ichiro Takemasa[5]

1 = Verily Life Sciences
2 = Google Japan
3 = Department of Surgery and Science, Graduate School of Medical Sciences, Kyushu University
4 = Sapporo Medical University School of Medicine, Department of Gastroenterology and Hepatology
5 = Department of Surgery, Surgical Oncology and Science, Sapporo Medical University
*authors share first authorship position
**authors share last authorship position

## Abstract

### Background and aims

Artificial Intelligence (AI) Computer-Aided Detection (CADe) is commonly used for polyp detection, but data seen in clinical settings can differ from model training. Few studies evaluate how well CADe detectors perform on colonoscopies from countries not seen during training, and none are able to evaluate performance without collecting expensive and time-intensive labels.

### Methods

We trained a CADe polyp detector on Israeli colonoscopy videos (5004 videos, 1106 hours) and evaluated on Japanese videos (354 videos, 128 hours) by measuring the True Positive Rate (TPR) versus false alarms per minute (FAPM). We introduce a colonoscopy dissimilarity measure called "MAsked mediCal Embedding Distance" (MACE) to quantify differences between colonoscopies, without labels. We evaluated CADe on all Japan videos and on those with the highest MACE.

### Results

MACE correctly quantifies that narrow-band imaging (NBI) and chromoendoscopy (CE) frames are less similar to Israel data than Japan whitelight (bootstrapped z-test, |z| > 690, $p < 10^{-8}$ for both). Despite differences in the data, CADe performance on Japan colonoscopies was non-inferior to Israel ones without additional training (TPR at 0.5 FAPM: 0.957 and 0.972 for Israel and Japan; TPR at 1.0 FAPM: 0.972 and 0.989 for Israel and Japan; superiority test t > 45.2, $p < 10^{-8}$). Despite not being trained on NBI or CE, TPR on those subsets were non-inferior to Japan overall (non-inferiority test t > 47.3, $p < 10^{-8}$, δ = 1.5% for both).

## Conclusion

Differences that prevent CADe detectors from performing well in non-medical settings do not degrade the performance of our AI CADe polyp detector when applied to data from a new country. MACE can help medical AI models internationalize by identifying the most "dissimilar" data on which to evaluate models.

*4-5 keywords*
Endoscopy; Computer-aided detection (CADe) system; Colorectal polyp; Detection; artificial intelligence

*DEI topics*
Parity

# Introduction

Colorectal cancer (CRC) is the second leading cause of cancer death globally [1], and results in an estimated 900,000 deaths each year [2]. Colonoscopy screening reduces mortality by identifying and removing polyps, which over time can become malignant [3]. However, the efficacy of colonoscopies is operator dependent, and 22% - 28% of polyps are missed due to reasons including fatigue, distraction and operator skill [4]. To address this concern, the use of computer aided detection (CADe) in colonoscopy is commonplace [5]. In particular artificial intelligence has proven especially effective, leading to trials and medical device applications globally [5].

A major challenge for artificial intelligence (AI) is generalizability between countries with widely differing healthcare systems [6-9]. This problem is particularly acute in colonoscopy, where hardware, scoping procedures, and differing imaging technologies raise important questions about how well AI will internationalize. In addition many public health and healthcare applications of AI require performance analyses on subsets of data, because topline metrics can hide patterns of errors, for example in protected subgroups [10-12]. Despite this, no studies have directly empirically measured how well an AI CADe polyp detector performs when used in an entirely different demographic from that on which it was trained.

Typically, AI models heavily rely on training data and generalize poorly to data that is different (called "domain adaptation") [13, 14]. The drop in performance can be significant [43]. In commercial applications of object detection, changes in style, resolution, and illumination usually require complicated techniques to adapt such as domain-specific classifiers and specialized loss functions [44], iteratively retraining the model on the new domain [45], or synthetic data and model retraining [46].

In this work, we explore how a CADe polyp detector trained on Israel colonoscopy videos performs on colonoscopy videos from Japan, two healthcare systems that differ widely across demographics, approaches to colonoscopy, and the technology itself. We evaluate the current published state of the art on video [15], measure performance, and demonstrate two important subsets of data where the detector performs better than might be expected. We demonstrate a new technique that quantifies which data subsets from a new country are most likely to be different from the training data, and can therefore be used to help polyp detector developers focus their time-consuming and expensive data collection efforts only on the subset most likely to be difficult.

# Materials and Methods

## Data collection and patients

**Japan**: This retrospective study was approved by the Ethics Committees of the Institutional Review Boards of Kyushu and Sapporo University Hospitals (Sapporo IRB: 322-157 Kyushu IRB: 2021-448). Colonoscopy videos were retrospectively collected from two large university hospitals in Kyushu (86 videos) and Sapporo (268 videos), Japan. Patients undergoing screening colonoscopies for any reason were eligible for inclusion. Patients under 18 years old, or who received colonoscopies for any other reason were excluded. Videos were recorded at various resolutions, formats, and containers. All data were de-identified; if personally identifiable information were displayed on the head-up display (HUD), the videos were re-encoded with the HUD obscured before being transferred out of the hospital.

**Israel**: The CADe polyp detector was trained and evaluated on 5004 videos from 3 hospitals in Israel. The videos were gathered from screening colonoscopy procedures. Each case consisted of a single video of an entire procedure (including the insertion phase) recorded at 30 frames per second, using a variety of endoscope models (see *Supplemental Section: Data*). All videos and metadata were deidentified, according to the Health Insurance Portability and Accountability Act Safe Harbor. Data from two hospitals were used to train the polyp detector, and data from the third was used for evaluation.

For details on data storage and preprocessing, see *Supplemental Section: Data Storage and Preprocessing.*

## Annotation procedure

The video annotation procedure is identical to [15] (section "Annotation procedure"). To summarize, each video was annotated by board-certified gastroenterologists from Israel and India, drawn from a pool of 20. Each video was labeled by two independent offline gastroenterologists who were not authors of the study. In cases where the two gastroenterologists disagreed, a third, more senior physician made a final decision based on the first two reads. All physicians used a specialized labeling tool to identify polyp bounding boxes. These annotations provide the ground truth for polyp locations. For more details, see *Supplemental Section: Data Annotation*.

## Neural network architecture

Details of the neural network architecture are provided in *Supplemental Section: Architecture of Neural Network Detector*. In summary, we use a standard single-shot detector (SSD) architecture RetinaNet [19] with publicly available hyperparameters [20]. The detector is trained only on data from Israeli hospitals. The detector takes a single video frame as input and outputs a (possibly empty) sequence of bounding boxes and the associated confidences. Note that the single-frame prediction model is the same as in [15], although we aggregate predictions across frames differently, as we discuss below.

Unlike the detector described in [15], we use the temporal filtering method used in [21], referred to here as Median Filtering. Instead of using a separate neural network to aggregate temporal information across neighboring frames, we use a median filter. This results in a "majority" vote across neighboring frames: for example, if the single-frame detector finds a polyp in a 7 frame window centered around the current frame, we consider the detector to have detected a polyp in this frame.

## Neural network training

The model was trained on data from Israel using standard stochastic gradient descent type techniques for minimizing detection loss, with the same procedure and parameters as [15]. The model was not fine tuned on data from Japan. For details on model training, please see *Supplemental Section: Neural Network Training.*

## Differences in Japan data

### Modifications to detector pipeline

While our study didn't change the CADe polyp detector neural network or filter, there were two important adjustments made to the video preprocessing between Israel and Japan. First, the HUD of the endoscopes used in the Japan colonoscopies were different than any seen in the Israel data, so the region-of-interest (ROI) on which the detector was run on was adjusted. Second, an inside-outside model [18] was used to detect which parts of the video were taken from inside the body, and which were outside. We used the inside-outside model to automatically remove the outside-of-the-body sections of the videos. Example frames from Israel and Japan colonoscopies can be seen in Figure 1.

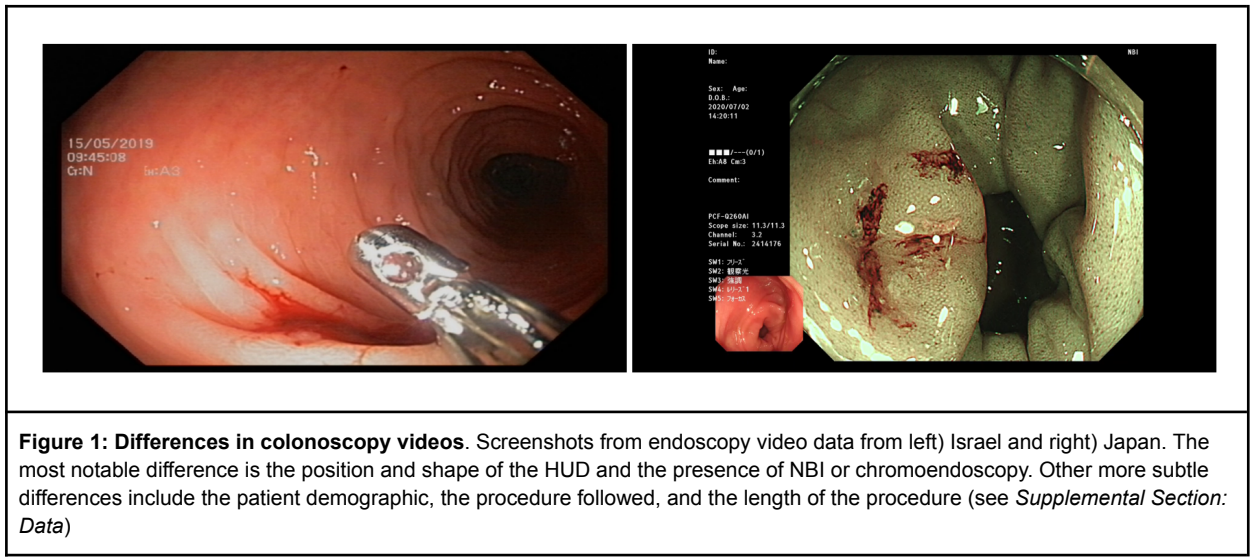

**Figure 1: Differences in colonoscopy videos**. Screenshots from endoscopy video data from left) Israel and right) Japan. The most notable difference is the position and shape of the HUD and the presence of NBI or chromoendoscopy. Other more subtle differences include the patient demographic, the procedure followed, and the length of the procedure (see *Supplemental Section: Data*)

## Narrow-band imaging and chromoendoscopy

Unlike the Israel dataset, parts of the Japanese videos contained narrow-band imaging (NBI), a technology that modifies the wavelength and bandwidth of a colonoscope's light to better visualize the colonic mucosa. To identify NBI, we extracted text in the HUD indicating whether NBI had been turned on. For statistics, we considered various fractions for the amount of time that a polyp was seen in NBI to be considered an NBI polyp (see Figure 3).

Furthermore, some of the frames with NBI also included chromoendoscopy (CE), a technique involving spraying dye to highlight edge contrasts [22]. To identify CE frames, we identified a range of pixel values that were present during CE and absent in a normal Japan colonoscopy. We detected the presence or absence of pixels in this range, and considered a frame to "have chromoendoscopy" if a sufficient fraction of pixels were in this range. We report performance on CE lesions for various values of the fraction of the lesions seen with CE (ex. 30% means 30% of the time the lesion is in view, it has CE).

## MAsked mediCal Embedding (MACE) Distance: Identifying dissimilar colonoscopy data using distributional similarity

In this section, we present a data-driven technique for quantifying differences in colonoscopy videos without manual annotations. As is the case for object detectors in other domains like facial recognition and identifying real-world objects, these techniques can be used to determine on which parts of a colonoscopy video the polyp detector is likely to perform poorly.

Our dissimilarity technique combines two technologies: representation learning and distributional distance. Representations are compact and specific encodings of the data, in our case images, that capture only the relevant information in a data-driven way. Distributional differences quantify the similarity between two collections of objects, in our case collections of representations of colonoscopy images, into a single number. Both are non trivial from a technological and mathematical perspective, and are discussed in greater detail in the following two paragraphs.

In our study, we use two styles of learning representations, Vision Transformers (ViT) [23] and Contrastive Learning (SimCLR) [25]. ViT is trained in a self-supervised way (no annotations required) by trying to predict "masked out" sections of the image during training, and generates representations useful for a number of other tasks [24]. Our model represents data in 384 dimensions. SimCLR learns representations by mapping images to representations that are "closer" in embedding space if they are "closer" in some meaningful way. In our SimCLR model, embeddings are trained to be closer if they are different views of the same polyp. Our SimCLR model was trained only on annotated Israel data, alleviating the need for Japan labels. It represents data in 128 dimensions. Representations learned from this proxy task have been used in a number of problems and modalities [26, 27]. When computing the MACE distance, we use only the cropped ROI (e.g. we remove the HUD). See Figure 2 for a schematic explanation of how SimCLR and ViT are trained.

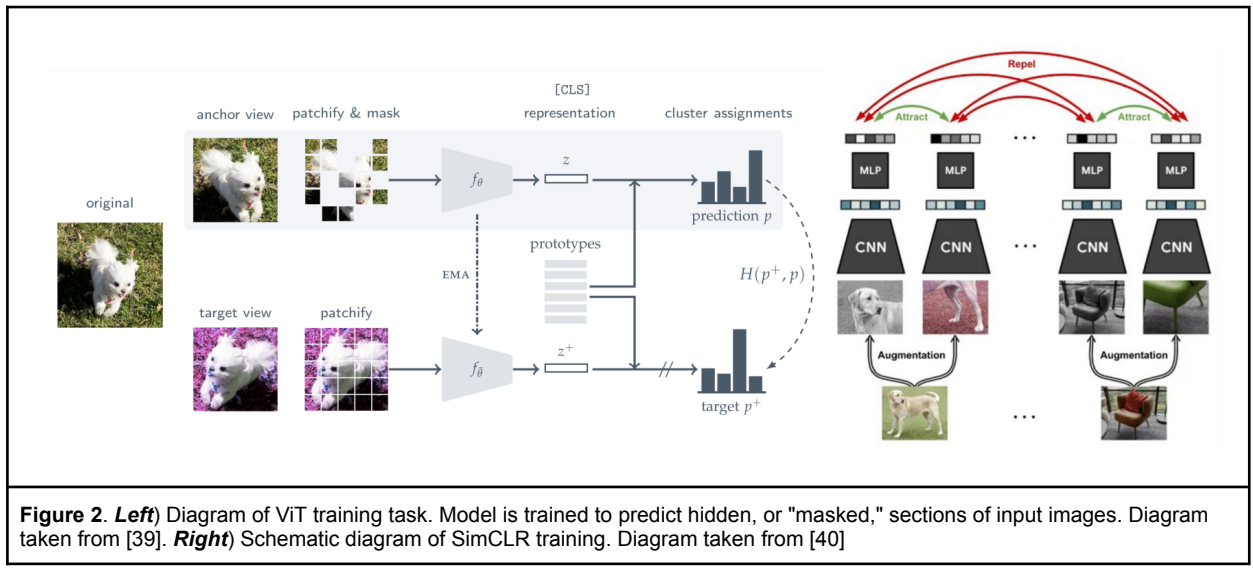

**Figure 2**. *Left*) Diagram of ViT training task. Model is trained to predict hidden, or "masked," sections of input images. Diagram taken from [39]. *Right*) Schematic diagram of SimCLR training. Diagram taken from [40]

We use the Frechet Distance as the definition of "closeness" in multidimensional representation space. This definition gives a notion of distance to collections of embeddings by fitting multivariate Gausian distributions to the data and measuring the difference in distribution space. This measure was re-popularized in the context of generative modeling [28], and we use the implementation in the popular TensorFlow Generative Adversarial Network library [29]. We use the Frechet distance instead of alternatives like Centered Kernel Alignment [41], which has also been used to compare embeddings [42], because of its strong theoretical foundations.

## Statistical analysis

To compare the performance of the CADe detector between Israel and Japan colonoscopy videos, we perform bootstrapped non-inferiority tests [47]. Specifically, we compare the False Alarm per Minute (FAPM) versus True Positive Rate (TPR) curves at two clinically-relevant operating points (FAPM at 0.5 and 1.0). We resample videos 1000 times. For each resample, we compute the FAPM vs TPR curve on both Israel and Japan. The model has a curve instead of a fixed point because the median filtering parameters can be adjusted to trade off between FAPM and TPR (see section *Neural network architecture* for details on the parameters, and the next paragraph for details on constructing the curve).We construct a 95% confidence interval on the difference between the two curves' TPRs and conduct a superiority test. We report the results of a one-sided test [47]. We compare curves by comparing TPR at fixed values of the false alarm rate, which is preferred in clinical contexts [30].

To analyze the performance of our system on the NBI and CE subsets compared to Japan whightlight, we compare the true positive rate of polyps on NBI against the TPR on all polyps. Specifically, for a given fraction, we only include polyps that are visible in NBI/CE at least that fraction of the time (number of such polyps given in *Supplemental Section: Data*). We compute the TPR in the TPR vs FAPM curve as the total number of detected polyps divided by the number of actual polyps on all videos. We compute the TPR at a given FAPM in the same way as for the Israel versus Japan case, by interpolating between neighboring points. For NBI and CE, videos without any NBI/CE polyps are excluded from the computation. We construct 95% confidence intervals on TPRs for our figures using 1000-resample bootstrapping, and perform bootstrapped superiority and non-inferiority tests on difference of means for statistical tests. We run tests at 0.5 FAPM and 1.0 FAPM for both CE vs WL and NBI vs WL, for visibility fractions at 10% intervals until the number of applicable polyps drops below 100 polyps. This methodology results in visibility fractions of 10-60% for both NBI and CE, and a total of 28 statistical tests. We use a confidence interval of 95% and a margin of 0.015 for all tests.

To determine the signifances of MACE differences between different subsets of videos (e.g. NBI vs WL as compared to CE vs WL), we perform a bootstrapped z-test on the MACE distances. We perform resampling on video subsets, which are usually polyps, rather than on entire videos, as we do in other bootstrapped tests. We resample 1000 times. Note that the MACE score is computed between two collections and is non-linear. We perform a two-sided test and report p-values and the test statistic.

# Results

Descriptives for the full datasets can be found in Table 1 and *Supplemental Section: Data*.

# Differences between Israel and Japan colonoscopy videos: Prevalence of NBI and CE

There are a number of differences between the colonoscopy videos collected from our Israel and Japan sites. The Japanese colonoscopy videos are longer (13 minutes as compared to 22 minutes, on average). There are more polyps per procedure in the Japan colonoscopy videos (0.7 in Israel as compared to 2.0 in Japan). In addition, NBI and CE are very commonly used in Japan colonoscopies. 9% of the data in Japan were viewed with NBI (11 hours total) as compared with 0% in Israel. 3% of the data from Japan were viewed with CE (3.8 hours total) as compared to 0% in Israel. For a summary of these differences, see Table 1.

|  | Israel train | Israel validation | Japan validation |
| --- | --- | --- | --- |
| Videos | 3611 | 1393 | 354 |
| Hours | 796 | 310 | 128 |
| Average video length (minutes) | 13 | 13 | 22 |
| Frames (million) | 86 | 33 | 14 |
| Unique polyps | 2230 | 956 | 721 |
| Polyps per procedure | 0.6 | 0.7 | 2.0 |
| NBI (hours) | 0 | 0 | 11 |
| Polyp seen in NBI (hours) | 0 | 0 | 4 |
| Chromoendoscopy (estimated, hours) | 0 | 0 | 3.8 |
| Polyps seen in chromoendoscopy (hours) | 0 | 0 | 0.8 |
| Nationality of patients | Israel-only | Israel-only | Japan-only |

**Table 1**: Data descriptives between Israel training, Israel validation, and Japan datasets. We highlight in red the clinically meaningful differences between the datasets.

NBI and CE use in Japan videos were substantial, both in terms of fraction of videos with them and amount of time lesions were viewed using one of these two techniques. Furthermore, NBI was used much more commonly than CE. Of the 353 Japan videos, 126 had NBI on for at least 5% of the video, and 16 videos had NBI on for at least 20%. 19 videos had CE for at least 5% of the video, and 4 videos had CE for at least 20%. For more details, see Figure 3 upper and Supplemental table 1.

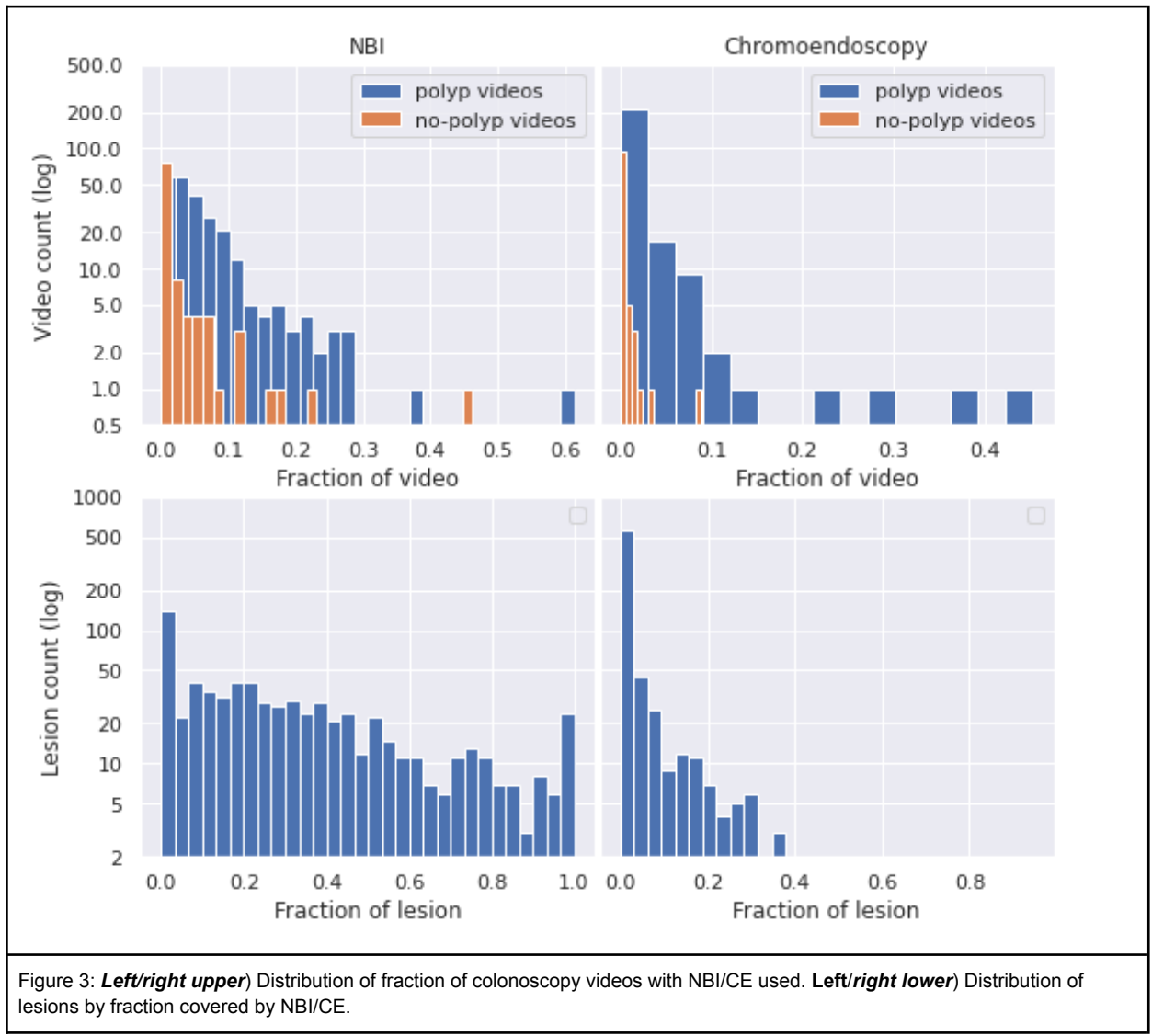

Figure 3: *Left/right upper*) Distribution of fraction of colonoscopy videos with NBI/CE used. *Left/right lower*) Distribution of lesions by fraction covered by NBI/CE.

Both NBI and CE were used significantly more frequently in polyp videos than videos without polyps. NBI was on for at least 5% of the video in 110/248 polyp videos, compared with 16/105 videos without polyps. Similarly, CE is used for at least 5% of the video in 18/248 polyp videos, compared with 1/105 videos without polyps. For more details, see Figure 3 upper and Supplemental table 1.

In addition to video coverage, polyps in the Japan colonoscopy videos are often viewed through NBI or CE. Of the 721 polyps in the Japan colonoscopy videos, 509/420 are seen with NBI/CE at least 10% of the time they are in view, 220/242 polyps are seen at least 40% of the time with NBI/CE, and 91/78 polyps are seen at least 70% of the time with NBI/CE. For more details, see Figure 3 lower and Supplemental Table 2.

## Data dissimilarity: MACE distances

Using ViT and SimCLR representations together with Frechet Distance, we find that Japan NBI and CE frames are consistently more dissimilar to the Israel training data than Japan white light (Table 2) using bootstrapped, one-sided difference of means z-test (t = -736.5, p < $10^{-8}$ for NBI and ViT, z = -676.4, p < $10^{-8}$ for NBI and SimCLR, z = -656.5, p < $10^{-8}$ for CE and ViT, z = -733.4, p < $10^{-8}$ for CE and SimCLR). We also see that CE frames are more dissimilar than NBI frames using bootstrapped, using two-sided difference of means z-test (t = -90.6, p < $10^{-8}$ for ViT, z = -87.7, p < $10^{-8}$ for ViT, z=, p < $10^{-8}$ for SimCLR). These difference orderings are the same whether we use MACE with ViT or MACE with SimCLR, supporting the claim that these differences are a result of the data and not an idiosyncrasy of the representation. In addition, MACE distances between polyp frames and non-polyp frames suggest that this metric might have some ability to detect polyp frames. See Table 2 for confidence intervals for the MACE differences between these groups.

|  | Embedding | |
| :---: | :---: | :---: |
| Comparison | ViT | SimCLR |
| Israel vs Japan WL | (779.3, 833.1) | (11.7, 12.8) |
| Israel vs Japan NBI | (1450.3, 1554.5) | (23.4, 25.3) |
| Israel vs Japan CE | (1559.8, 1705.9) | (25.4, 27.5) |
| Israel polyp vs Israel no polyp | (549.8, 594.6) | (6.2, 6.8) |
| Israel polyp vs Japan polyp | (1193.2, 1217.6) | (12.4, 13.4) |

**Table 2: MACE distance 95% confidence intervals, bootstrapped. 1000 bootstrapping resamples.**

In addition to the quantitative analysis above, we visualize the 384-d ViT embeddings and the 128-d SimCLR embeddings in 2D using T-SNE [31] (see Figure 4). Note that T-SNE generates clusters without label information. T-SNE clusters both embeddings into groups with understandable meanings, giving visual evidence that these embeddings contain semantically meaningful information, and that they are easily accessible to data-driven algorithms. These clusters further support the usefulness of MACE as an annotation-free method of quantifying data differences in a way which aligns with human intuition.

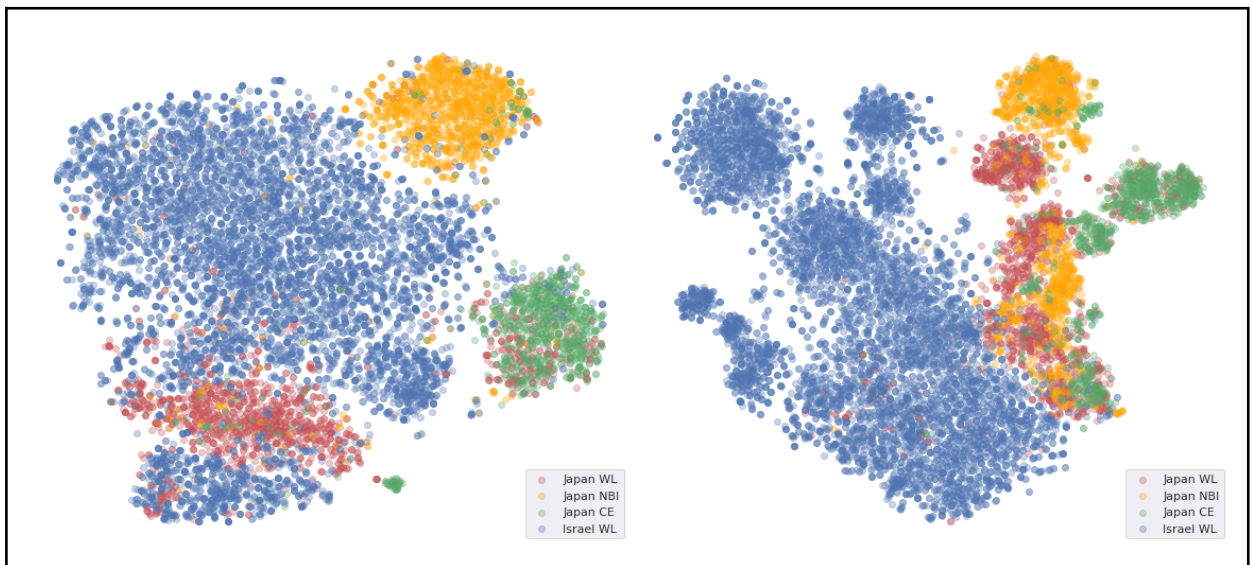

**Figure 4:** T-SNE [31] clustering on left) SimCLR [25], right) ViT [24] embeddings, between Israel WL, Japan WL, Japan NBI, Japan CE frames. We see that both embeddings are able to distinguish between frame types, without costly and time-intensive annotations on the Japan data. Note that ViT (right) doesn't use annotations of any kind, while SimCLR (left) uses polyp annotations from the Israel data only.

## CADe Polyp detector performance

The overall performance of the CADe polyp detector is non-inferior on Japan as compared to unseen Israel data, despite only being trained on Israel colonoscopy data (superiority test t > 45.2, $p < 10^{-8}$). First, we see that at two clinically-relevant operating points, determined by consulting with clinical experts, the performance is not worse. At 0.5 false alarms per minute, the models have a TPR of 0.957 / 0.972 for Israel / Japan, and at 1.0 false alarms per minute a TPR of 0.972 / 0.989 for Israel / Japan (see Figure 5 left). Importantly, we also find that there are no clinically relevant operating points where the model performs substantially worse on Japan colonoscopy vidoes than on Israel ones.

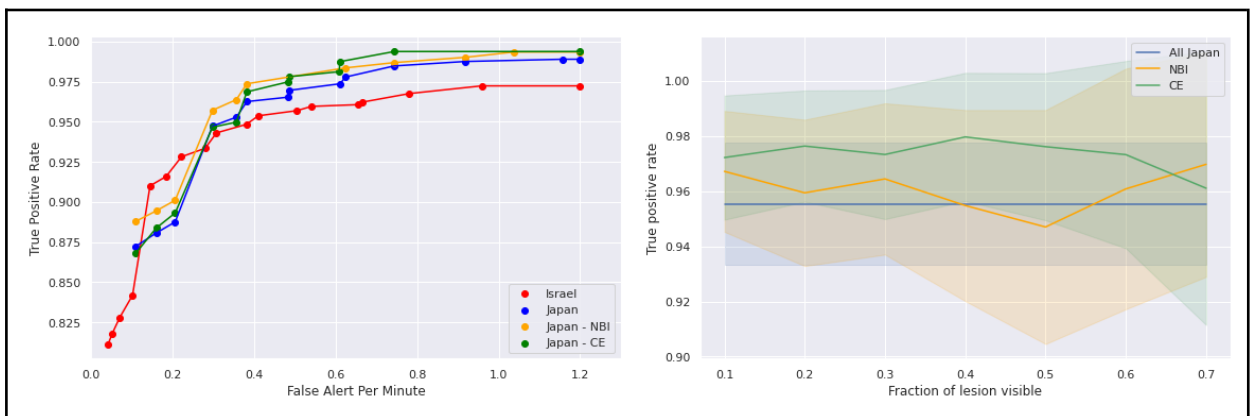

***Figure 5: Left*)** True positive rate (TPR) vs false alarm rate of CADe polyp detector on Israel evaluation colonoscopy videos, all Japan videos, only NBI polyps, and only CE polyps. NBI/CE lesions are those that at least 30% of the time are visible in NBI/CE.

303 lesions of the 721 in the Japan dataset are visible in NBI 30% or more, 318 are visible in chromoendoscopy (see . **Right**) Impact of NBI and CE on TPR. Each point only looks at lesions that are seen with NBI/CE at least that fraction of the time. For this graph, we look at the best parameters that give 0.5 FAPM. For the same graph at 1.0 FAPM, see Supplemental section "True Positive Rate Analysis".

Second, we found that the polyp detector TPR is non-inferior on NBI and CE as compared to Japan whitelight (non-inferiority test t > 47.3, p < $10^{-8}$, δ = 1.5% for both). Importantly, we find that the detector performance is non-inferior across all values that we measured of the fraction of the polyp seen under NBI/CE, which are 0.1 increments until the number of lesions falls below 100 (see Figure 5 right for the TPR values at 0.5 FAPM. See Supplemental Figure 3 for a TPR comparison at 1.0 FAPM, and Supplemental Table 3 for confidence intervals). This is surprising, since the model was not trained on any data using NBI or CE.

# Discussion

This paper makes two main contributions. First, we explore how changes in colonoscopy procedures affect the performance of AI CADe polyp detectors. The changes we explore are those found between colonoscopy procedures done in Israel and Japan, but are representative of changes that can be found between any two countries (demographics, approaches to colonoscopy, endoscope technology, etc). In order to quantify the dissimilarity in the colonoscopy procedures between Japan and Israel, and to identify the most dissimilar parts of the procedures, we introduce MACE, a novel method for quantifying the dissimilarity between training and evaluation data in medical imaging. We demonstrate the ability of our approach to reflect clinically important differences between the two datasets. Second,, we show that despite large differences in data distribution, our AI system for colonoscopy generalizes well across two different healthcare systems, maintaining performance with clinically applicable sensitivity and false alarm rates. Our approach makes it possible to use data-driven methods to quantify differences in data that might present challenges to model generalizability, and we demonstrate that CADe polyp detectors perform better than expected on novel data.

MACE also has potential use in selecting data for model development and evaluation. Researchers can use our approach to identify data where the MACE distance is higher, and preferentially collect more of this data in order to both improve model performance, and to more fully explore model performance ahead of clinical studies in order to reduce the risk a model performs poorly due to differences in data distribution.

Our model may have a training role in supporting human gastroenterologists when encountering the new imaging modality for the first time. The rapid generalizability of our approach to new modalities, and the aforementioned learning curve for trainees, suggests our model may be useful in providing AI augmented education to less experienced practitioners. This approach is already being adopted in radiology training programs [34, 35], with evidence that it can improve radiologist performance [36]. Our approach may be particularly suitable to continued medical education, especially when new endoscopy modalities are introduced and many clinicians will experience a learning curve while they adapt to the new technology.

This study has a number of limitations. First, we consider only two countries, Israel and Japan. Practice can vary even site to site [33], and future work should demonstrate the approach on a greater number of countries and sites. Second, our novel approach to colonoscopy dissimilarity is one of many, and a thorough investigation of all potential approaches is out of scope for this publication. Third, NBI and CE are more commonly used when polyps are seen or expected, making detector performance during these techniques less important than during whitelight for finding new polyps. Fourth, our work is not prospective. Finally, generalizability is only one factor that impacts the use of AI in colonoscopy. Other factors like variation in user ability [38], or differences in operational environment [32] can create practical challenges to the use of AI independent of the model performance itself.

In conclusion, we demonstrate the generalizability of a real time polyp detection AI model between Israel and Japan, including to previously unseen NBI videos. We show that initial lower performance in Japanese CE videos can be predicted, and propose an approach to maintain original performance by training on a very small number of additional examples. Our results enable generalizability to new locations and imaging modalities, and improve the feasibility of using AI in colonoscopy.

# Acknowledgements

We would like to thank Anil Patwardhan and Carson McNeil for their insightful and incredibly helpful feedback on our manuscript. We'd like to thank Mira Fruchter for her help organizing the data collection efforts. We'd like to thank Karen Ouk for her organizational support on this project, and we'd like to thank Daniel Freedman, Jonathan Huang, and Vighnesh Birodkar for their technical assistance. We would like to thank Verily Life Sciences for funding this work.

# References

1. Marianne Grønlie Guren, The global challenge of colorectal cancer, The Lancet Gastroenterology & Hepatology, Volume 4, Issue 12, 2019, Pages 894-895, ISSN 2468-1253, https://doi.org/10.1016/S2468-1253(19)30329-2.
2. Rawla P, Sunkara T, Barsouk A. Epidemiology of colorectal cancer: incidence, mortality, survival, and risk factors. Prz Gastroenterol. 2019;14(2):89-103. doi: 10.5114/pg.2018.81072. Epub 2019 Jan 6. PMID: 31616522; PMCID: PMC6791134.
3. Zauber AG, Winawer SJ, O'Brien MJ, Lansdorp-Vogelaar I, van Ballegooijen M, Hankey BF, Shi W, Bond JH, Schapiro M, Panish JF, Stewart ET, Waye JD. Colonoscopic polypectomy and long-term prevention of colorectal-cancer deaths. N Engl J Med. 2012

Feb 23;366(8):687-96. doi: 10.1056/NEJMoa1100370. PMID: 22356322; PMCID: PMC3322371.
4. Leufkens AM, van Oijen MG, Vleggaar FP, Siersema PD. Factors influencing the miss rate of polyps in a back-to-back colonoscopy study. Endoscopy. 2012 May;44(5):470-5. doi: 10.1055/s-0031-1291666. Epub 2012 Mar 22. PMID: 22441756.
5. Mori Y, Kudo SE, Berzin TM, Misawa M, Takeda K. Computer-aided diagnosis for colonoscopy. Endoscopy. 2017 Aug;49(8):813-819. doi: 10.1055/s-0043-109430. Epub 2017 May 24. PMID: 28561195; PMCID: PMC6193286.
6. Norori N, Hu Q, Aellen FM, Faraci FD, Tzovara A. Addressing bias in big data and AI for health care: A call for open science. Patterns (N Y). 2021 Oct 8;2(10):100347. doi: 10.1016/j.patter.2021.100347. PMID: 34693373; PMCID: PMC8515002.
7. De Fauw J, Ledsam JR, Romera-Paredes B, et al. Clinically applicable deep learning for diagnosis and referral in retinal disease. Nature Medicine. 2018 Sep;24(9):1342-1350. DOI: 10.1038/s41591-018-0107-6. PMID: 30104768.
8. Winkens, Jim et al. "Contrastive Training for Improved Out-of-Distribution Detection." ArXiv abs/2007.05566 (2020): n. Pag.
9. Shor J, Miyatani Y, Arita E, Chen P, Ito Y, Kayama H, Reiter J, Kobayashi K, Kobayashi T, Reducing Health Anxiety in Patients With Inflammatory Bowel Disease Using Video Testimonials: Pilot Assessment of a Video Intervention, JMIR Form Res 2023;7:e39945
10. Joy Buolamwini, Timnit Gebru Proceedings of the 1st Conference on Fairness, Accountability and Transparency, PMLR 81:77-91, 2018.
11. Moritz Hardt, Eric Price, Nati Srebro, et al. 2016. Equality of opportunity in supervised learning. In Advances in neural information processing systems. 3315–3323.
12. Arık, S.Ö., Shor, J., Sinha, R. et al. A prospective evaluation of AI-augmented epidemiology to forecast COVID-19 in the USA and Japan. npj Digit. Med. 4, 146 (2021). https://doi.org/10.1038/s41746-021-00511-7
13. Motiian, S., Piccirilli, M., Adjeroh, D., & Doretto, G. (2017). Unified Deep Supervised Domain Adaptation and Generalization. In Proceedings of the IEEE International Conference on Computer Vision (ICCV).
14. J. Shor et al. 2019. Personalizing ASR for Dysarthric and Accented Speech with Limited Data. In Proc. Interspeech 2019, pages 784–788.
15. D. Livovsky et al, Detection of elusive polyps using a large-scale artificial intelligence system (with videos), Gastrointestinal Endoscopy, Volume 94, Issue 6, 2021, Pages 1099-1109.e10, ISSN 0016-5107, https://doi.org/10.1016/j.gie.2021.06.021.
16. J. Shor & N. Johnston (2023). DOES VIDEO COMPRESSION AFFECT CADE POLYP DETECTORS? In Gastrointestinal Endoscopy (Vol. 97, Issue 6, p. AB768). Elsevier BV. https://doi.org/10.1016/j.gie.2023.04.1256
17. J. Shor and N. Johnston, "The Need for Medically Aware Video Compression in Gastroenterology", in Neural Information Systems Processing, Workshop on Medical Imaging meets NeurIPS, 2022.
18. O. Kelner et al, "Semantic Parsing of Colonoscopy Videos with Multi-Label Temporal Networks", arXiv e-prints, 2023. doi:10.48550/arXiv.2306.06960.

19. T. -Y. Lin, P. Goyal, R. Girshick, K. He and P. Dollár, "Focal Loss for Dense Object Detection," in IEEE Transactions on Pattern Analysis and Machine Intelligence, vol. 42, no. 2, pp. 318-327, 1 Feb. 2020, doi: 10.1109/TPAMI.2018.2858826.
20. Google research. (2020, July 10). object detection config. GitHub. https://github.com/tensorflow/models/blob/master/research/object_detection/configs/tf2/ssd_resnet50_v1_fpn_640x640_coco17_tpu-8.config
21. C. Leggett et al. (2023). Interim Results from the Gastroenterology Artificial Intelligence System for Detecting Colorectal Polyps (GAIN) Trial: An International Community-Based Randomized Controlled Multi-Center Trial.
22. ASGE Technology Committee; Wong Kee Song LM, Adler DG, Chand B, Conway JD, Croffie JM, Disario JA, Mishkin DS, Shah RJ, Somogyi L, Tierney WM, Petersen BT. Chromoendoscopy. Gastrointest Endosc. 2007 Oct;66(4):639-49. doi: 10.1016/j.gie.2007.05.029. Epub 2007 Jul 23. PMID: 17643437.
23. A. Dosovitskiy, et al, (2021). An Image is Worth 16x16 Words: Transformers for Image Recognition at Scale. In International Conference on Learning Representations.
24. L. Yuan et al, (2021). Tokens-to-Token ViT: Training Vision Transformers From Scratch on ImageNet. In Proceedings of the IEEE/CVF International Conference on Computer Vision (ICCV) (pp. 558-567).
25. T. Chen, et al, A Simple Framework for Contrastive Learning of Visual Representations. Proceedings of the 37th International Conference on Machine Learning, PMLR 119:1597-1607, 2020.
26. R. Qian, et al., "Spatiotemporal Contrastive Video Representation Learning," in 2021 IEEE/CVF Conference on Computer Vision and Pattern Recognition (CVPR), Nashville, TN, USA, 2021 pp. 6960-6970. doi: 10.1109/CVPR46437.2021.00689
27. J. Shor et al (2020) Towards Learning a Universal Non-Semantic Representation of Speech. Proc. Interspeech 2020, 140-144, doi: 10.21437/Interspeech.2020-1242
28. M. Heusel et al 2017. GANs trained by a two time-scale update rule converge to a local nash equilibrium. In Proceedings of the 31st International Conference on Neural Information Processing Systems (NIPS'17). Curran Associates Inc., Red Hook, NY, USA, 6629–6640.
29. J. Shor and S. Cotado. "Computing Systems with Modularized Infrastructure for Training Generative Adversarial Networks." U.S. Patent No. 11,710,300. 12 Oct. 2018.
30. Pepe M, Longton G, Janes H. Estimation and Comparison of Receiver Operating Characteristic Curves. Stata J. 2009 Mar 1;9(1):1. PMID: 20161343; PMCID: PMC2774909.
31. van der Maaten, L. & Hinton, G. (2008). Visualizing Data using t-SNE . Journal of Machine Learning Research, 9, 2579--2605.
32. Li JW, Ang TL. Colonoscopy and artificial intelligence: Bridging the gap or a gap needing to be bridged? Artif Intell Gastrointest Endosc 2021; 2(2): 36-49
33. Day LW, Bhuket T, Inadomi JM, Yee HF. Diversity of endoscopy center operations and practice variation across California's safety-net hospital system: a statewide survey. BMC Res Notes. 2013 Jun 15;6:233. doi: 10.1186/1756-0500-6-233. PMID: 23767938; PMCID: PMC3693938.

34. Shukla R, Salem M, Hou JK. Use and barriers to chromoendoscopy for dysplasia surveillance in inflammatory bowel disease. World J Gastrointest Endosc. 2017 Aug 16;9(8):359-367. doi: 10.4253/wjge.v9.i8.359. PMID: 28874956; PMCID: PMC5565501.
35. Tejani AS, Elhalawani H, Moy L, Kohli M, Kahn CE Jr. Artificial Intelligence and Radiology Education. Radiol Artif Intell. 2022 Nov 16;5(1):e220084. doi: 10.1148/ryai.220084. PMID: 36721409; PMCID: PMC9885376.
36. Fischetti C, Bhatter P, Frisch E, Sidhu A, Helmy M, Lungren M, Duhaime E. The Evolving Importance of Artificial Intelligence and Radiology in Medical Trainee Education. Acad Radiol. 2022 May;29 Suppl 5:S70-S75. doi: 10.1016/j.acra.2021.03.023. Epub 2021 May 18. PMID: 34020872.
37. Cheng, CT., Chen, CC., Fu, CY. et al. Artificial intelligence-based education assists medical students' interpretation of hip fracture. Insights Imaging 11, 119 (2020). https://doi.org/10.1186/s13244-020-00932-0
38. Sano Y, Ikematsu H, Fu KI, Emura F, Katagiri A, Horimatsu T, Kaneko K, Soetikno R, Yoshida S. Meshed capillary vessels by use of narrow-band imaging for differential diagnosis of small colorectal polyps. Gastrointest Endosc. 2009 Feb;69(2):278-83. doi: 10.1016/j.gie.2008.04.066. Epub 2008 Oct 25. PMID: 18951131.
39. Assran, Caron, Misra, Bojanowski, Bordes, F., Vincent, Joulin, Rabbat, & Ballas, N. (2022). Masked Siamese Networks for Label-Efficient Learning. arXiv preprint arXiv:2204.07141.
40. T. Chen, G. Hinton. "Advancing Self-Supervised and Semi-Supervised Learning with SimCLR." Google research blog, Apr 8 2020, https://ai.googleblog.com/2020/04/advancing-self-supervised-and-semi.html
41. S. Kornblith et al., "Similarity of neural network representations revisited," in ICML. PMLR, 2019, pp. 3519–3529.
42. J. Shor et al, "Universal Paralinguistic Speech Representations Using self-Supervised Conformers," *ICASSP 2022 - 2022 IEEE International Conference on Acoustics, Speech and Signal Processing (ICASSP)*, Singapore, Singapore, 2022, pp. 3169-3173, doi: 10.1109/ICASSP43922.2022.9747197.
43. R. Gopalan, Ruonan Li and R. Chellappa, "Domain adaptation for object recognition: An unsupervised approach," *2011 International Conference on Computer Vision*, Barcelona, Spain, 2011, pp. 999-1006, doi: 10.1109/ICCV.2011.6126344.
44. Chen, Y., Li, W., Sakaridis, C., Dai, D., & Van Gool, L. (2018). Domain adaptive faster r-cnn for object detection in the wild. In Proceedings of the IEEE conference on computer vision and pattern recognition (pp. 3339-3348).
45. J. Xu, S. Ramos, D. Vázquez and A. M. López, "Domain Adaptation of Deformable Part-Based Models," in *IEEE Transactions on Pattern Analysis and Machine Intelligence*, vol. 36, no. 12, pp. 2367-2380, 1 Dec. 2014, doi: 10.1109/TPAMI.2014.2327973.
46. Inoue, N., Furuta, R., Yamasaki, T., & Aizawa, K. (2018). Cross-domain weakly-supervised object detection through progressive domain adaptation. In Proceedings of the IEEE conference on computer vision and pattern recognition (pp. 5001-5009).

47. Walker E, Nowacki AS. Understanding equivalence and noninferiority testing. J Gen Intern Med. 2011 Feb;26(2):192-6. doi: 10.1007/s11606-010-1513-8. Epub 2010 Sep 21. PMID: 20857339; PMCID: PMC3019319.